\pdfoutput=1
\documentclass[11pt]{article}

\usepackage[utf8]{inputenc}
\usepackage[T1]{fontenc}
\usepackage{lmodern}
\usepackage[margin=1in]{geometry}
\usepackage{microtype}
\usepackage{graphicx}
\usepackage{booktabs}
\usepackage[font=small,labelfont=bf,skip=5pt]{caption}
\usepackage{amsmath, amssymb}
\usepackage{xcolor}
\usepackage{natbib}
\usepackage{hyperref}
\usepackage{cleveref}

\hypersetup{colorlinks=true, linkcolor=blue!55!black, citecolor=blue!55!black,
  urlcolor=blue!55!black}

\graphicspath{{figures/}}

\newif\ifanonymous
\anonymousfalse   

\ifanonymous
  \title{\textbf{Character Information in Language Models:\\
    A Prompt-Independent Word Store and a Late, Question-Driven Signal That Grows With Scale}}
  \author{Anonymous Authors}
\else
  \title{\textbf{Character Information in Language Models:\\
    A Prompt-Independent Word Store and a Late, Question-Driven Signal That Grows With Scale}%
    \thanks{Code and data: \url{https://github.com/pratyush-acharya/char-store-probing}}}
  \author{%
    Pratyush Acharya$^{1}$\thanks{Equal contribution.}\kern0.4em
    \thanks{Corresponding author: \texttt{acharya.pratyush12@gmail.com}}\kern0.4em\quad
    Nuraj Rimal$^{2}$\footnotemark[2]\kern0.4em\quad
    Apar Baral$^{3}$\\
    Habish Dhakal$^{1}$\quad
    Piyush Mohan Bhattarai$^{4}$\\[6pt]
    \normalsize $^{1}$SecurityPal AI\quad
    $^{2}$Southern Illinois University Edwardsville\\
    \normalsize $^{3}$University of Maryland\quad
    $^{4}$Saint Louis University
  }
  \hypersetup{pdftitle={Character Information in Language Models: A Prompt-Independent Word Store
      and a Late, Question-Driven Signal That Grows With Scale},
    pdfauthor={Pratyush Acharya, Nuraj Rimal, Apar Baral, Habish Dhakal, Piyush Mohan Bhattarai}}
\fi
\date{\today}

\begin{document}
\maketitle

\begin{abstract}
Linear probes recover character properties of a word from a language model's hidden states, but a
probe at the answer position cannot say whether the property was stored with the word or produced
because the prompt asked for it. We separate the two with a word-position baseline: the same
held-out-word probe applied to the query word's own tokens in a sentence that asks nothing. Across
16 open-weight models from seven families (0.6B--14B) and eleven character tasks, the word tokens
encode \texttt{first\_letter}, \texttt{last\_letter} and \texttt{most\_frequent\_letter} above a
word-only nuisance floor in all 16 models, and about 70\% of that signal is already present in the
token embedding. Asking the question adds a second signal at the answer position. Its size grows
with model scale for all four prompt-sensitive tasks (Holm-corrected permutation $p \leq 0.018$,
also within model families), and the growth is confined to the late layers: restricted to the
first quarter of the network, the slope is indistinguishable from zero. Small models' answer
positions hold less linearly decodable signal than their word-level store; large models' hold more. Four controls bound
these claims. Measured against the nuisance floor, the question-free signal that a majority-baseline
analysis counts as retained largely disappears (\texttt{last\_letter}: $-0.01$). Count tasks
become measurable only after length, repetition and query-letter shortcuts are removed by design.
A pre-registered test finds that a question placed before the word does not change the word's own
encoding beyond what an arbitrary prefix change does. Representation--readout gaps show no robust
scaling once probe saturation is accounted for. Pre-registered criteria, a fixed estimator
(held-out layer selection, cold-start folds) and all analysis code accompany the paper.
\end{abstract}

\section{Introduction}
\label{sec:intro}

Linear probes often recover a correct answer from a language model's hidden states when the
model's own output is wrong. This representation--output gap has been studied for factual and
character-level tasks \citep{kappa,early_encode,inside_out}, and it is often read as
evidence that the model has the answer but fails to express it. For character properties that
reading skips a step, because the last letter of a word is a fixed property of the word and
subword embeddings are known to encode spelling \citep{kaushal2022tokens,itzhak2022spelling}. A
probe at the answer position may therefore read a property that was stored with the word, one
that the question caused the model to assemble, or both.

\begin{figure}[t]
  \centering
  \includegraphics[width=\linewidth]{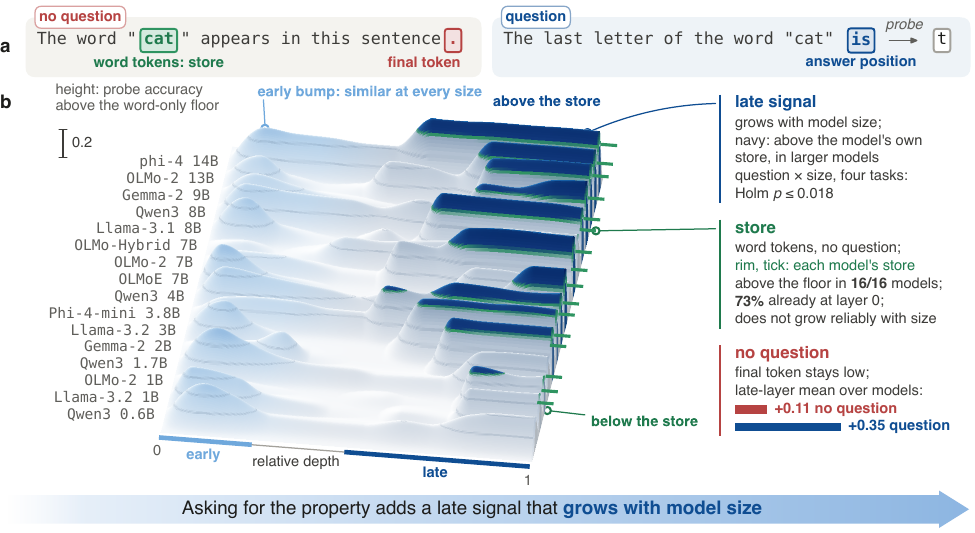}
  \caption{Overview. (a)~One held-out-word linear probe (here \texttt{last\_letter}) is read at the
    query word's own tokens in a sentence that asks nothing (green: the word-level store), at that
    sentence's final token (red), and at the answer position of the question prompt (blue).
    (b)~Each mountain is one model (16, smallest in front); its crest is probe accuracy above the
    word-only floor at the answer position under the question, across relative depth (zero shots).
    Asking adds an early bump near 10\% depth that is similar at every size and a late signal past
    half depth that grows with size. Navy marks where a mountain rises above its own store (green
    rim and tick), that is, where the answer position is more linearly decodable than the word's
    tokens; five models never do. The store does not grow reliably with size
    (\S\ref{sec:store}, \S\ref{sec:question}). Construction details are in
    Appendix~\ref{app:fig1}.}
  \label{fig:overview}
\end{figure}

We separate these with a \emph{word-position baseline}: the same held-out-word probe, applied to
the query word's own tokens inside a sentence that asks nothing
(\texttt{The word "cat" appears in this sentence.}). Under causal attention those tokens have seen
only \texttt{The word "}, so their decodability is the closest available measure of what the model
stores about the word independently of any task. We compare it with the answer position under
the task question and under two question-free carriers, for 16 open-weight models from seven
families (0.6B--14B) and eleven character tasks, using held-out-word cross-validation, a word-only
nuisance floor, held-out layer selection, and pre-registered decision criteria.

\paragraph{A prompt-independent store exists, and it is mostly lexical.}
The word tokens encode \texttt{first\_letter}, \texttt{last\_letter} and
\texttt{most\_frequent\_letter} above the nuisance floor in all 16 models
(\S\ref{sec:store}). Roughly 70\% of that above-floor signal is already present at layer 0, in the
token embedding, consistent with prior work on spelling in embeddings. This half of the picture is
confirmation, not discovery; it supplies the baseline that the rest of the analysis needs.

\paragraph{Asking adds a late signal, and only that signal grows with scale.}
At the answer position, the difference between the question prompt and the question-free carrier
grows with log parameter count for all four prompt-sensitive tasks, after Holm correction and
within model families (\S\ref{sec:question}; Figure~\ref{fig:twophase}b). Restricting the probe to layers past half depth
leaves the effect intact; restricting it to the first quarter removes it. The answer position of
small models holds less linearly decodable signal than their word-level store, while that of large
models holds more
(Figure~\ref{fig:overview}). Layer
profiles show the two phases directly (Figure~\ref{fig:twophase}a): an early bump that is similar
at every scale, then a late rise that is much stronger in larger models.

\paragraph{Controls change what can be claimed.}
Measured against the word-only floor, most of the question-free signal that a majority-baseline
analysis counts as retained is nuisance
(\S\ref{sec:question}). Count tasks fail the floor until length, letter repetition and query-letter
identity are removed by design (\S\ref{sec:nuisance}). Representation--readout gaps do not follow
a robust scaling trend once probe saturation is accounted for (\S\ref{sec:readout}). A
pre-registered stem-order experiment finds that placing the question before the word does not
change the word's encoding beyond what an arbitrary two-word prefix change does
(Appendix~\ref{app:prereg}).

\paragraph{Contributions.}
(1) A word-position baseline that separates prompt-independent storage from question-conditioned
decodability at the answer position. (2) Evidence that the scale-dependent part of character
decodability is late and question-driven, while the early signal and the word-level store do not grow reliably with scale.
(3) A set of controls (word-only floors with designed-out shortcuts, held-out layer selection,
pre-registration) that each reversed or removed a claim an uncontrolled analysis would have made.
We do not identify the mechanism behind the late signal; whether it reflects computation,
integration across the word's tokens, or reformatting is left open.

\section{Related Work}
\label{sec:related}

\paragraph{Character information in subword models.}
Token embeddings of pretrained models encode much of their tokens' character composition, and this
improves with model size \citep{kaushal2022tokens}; the embedding layers of RoBERTa and GPT-2
suffice to spell up to a third of the vocabulary \citep{itzhak2022spelling}. Layer-wise probes find
that letters beyond the first are only partly present in the embedding and are reconstructed in
intermediate and upper layers \citep{hiraoka_inui}. Controlled training comparisons
attribute the acquisition of such knowledge to tokenization-related factors (merge rules,
orthographic constraints) and to semantic and syntactic factors independent of tokenization
\citep{sato2026acquire}. Character tasks nonetheless remain hard at the output: the Strawberry
Problem traces the emergence of character-level performance during training \citep{strawberry},
CUTE finds that models can spell their tokens but often fail to use that knowledge to manipulate
text \citep{edman2024cute}, CharBench identifies word length and character count as predictors of
errors \citep{charbench}, and stochastic segmentation improves fine-grained subword processing
\citep{stochastok}. For factual recall, attributes are enriched at the subject's last token and
later extracted to the final position according to the relation \citep{geva2023dissecting}. Our
word-position baseline measures the analogous two sites for character properties, by
decodability rather than by intervention: what is stored at the word in context, and what the
question adds at the answer position.

\paragraph{Representation--output discrepancies.}
Linear or diagnostic readouts can recover information a model does not express, for
multiple-choice questions \citep{kappa} and hidden factual knowledge \citep{inside_out}.
Mechanistic studies of counting report early
encoding followed by late suppression \citep{early_encode} and a format-sensitive late MLP block
that overwrites a probe-correct repeated-token count \citep{repeated_token}. For cultural
knowledge, probe--output discrepancies decrease with scale but persist within families
\citep{chelombitko2026cultural}. We find that character-level discrepancies do not share a single
scaling trend once saturation is accounted for.

\paragraph{Probe interpretation.}
Probe accuracy depends on the representation, probe class, data and split; control tasks and
selectivity are therefore central \citep{hewitt2019control,belinkov2022probing}. Vocabulary overlap
permits word-level memorization \citep{rosa2020memorization,dhar2020vocab}, and the amount of
probe data changes conclusions \citep{voita2020mdl,zhu2022data}. An input-only baseline can
nearly match a hidden-state probe on arithmetic \citep{grewal2026decodability}, and removing format
nuisances can reduce a perfectly separating probe to chance \citep{sahoo2026format}.
\citet{early_encode} omit control
tasks on the grounds that their dataset construction prevents memorization; we do not rely on
that argument and use grouped splits by query word and word-only nuisance floors instead. We do not run a random-label selectivity control.

\paragraph{Prompts, demonstrations and readout geometry.}
Demonstrations convey label space and format even with degraded input--label mappings
\citep{min2022rethinking}; task recognition and task learning can be separated
\citep{pan2023what}, and scale changes in-context behavior \citep{wei2023larger}. Our zero-shot
contrasts remove demonstrations from every arm. Embedding-space analyses and tuned lenses relate
hidden states to the output vocabulary \citep{dar2023embedding,belrose2023tunedlens}, and amnesic
probing removes a learned subspace \citep{elazar2021amnesic}. A counting study finds a count
direction nearly orthogonal to each digit-token row \citep{garcia2026direction}; we test the joint
span of answer-token rows, a different object. Inference-time interventions can improve selected
outputs \citep{iti}; we make no causal claim.

\section{Experimental Setup}
\label{sec:setup}

\paragraph{Tasks.}
Nine character tasks over English words of length 3--9: \texttt{first\_letter},
\texttt{last\_letter}, \texttt{contains\_letter} (yes/no), \texttt{extraction} ($k$-th letter),
\texttt{counting} (occurrences of a queried letter), \texttt{most\_frequent\_letter},
\texttt{letter\_before} (the letter preceding a queried letter), \texttt{word\_length} and
\texttt{count\_distinct}. Two further tasks, \texttt{counting\_lm} and \texttt{count\_distinct\_lm},
ask the same questions as the two count tasks but sample examples so that word-only shortcuts
carry no information about the answer (\S\ref{sec:nuisance}, Appendix~\ref{app:lm}). Five tasks
(\texttt{first\_letter}, \texttt{last\_letter}, \texttt{most\_frequent\_letter},
\texttt{word\_length}, \texttt{count\_distinct}) are pure functions of the word; the others carry a
query letter or position. Prompts and answer sets are in Appendix~\ref{app:prompts}.

\paragraph{Stimuli and models.}
Each task has 800 examples sampled from a 6000-word frequency-ranked English vocabulary built from
the Brown Corpus \citep{kucera1967brown}, covering 405--757 unique words per task. We evaluate 16
open-weight models from seven families: Qwen3 (0.6/1.7/4/8B), Llama-3.x (1/3/8B), Gemma-2
(2/9B), OLMo-2 (1/7/13B), Phi-4 (3.8B-mini, 14B), OLMo-Hybrid-7B and OLMoE-1B-7B
\citep{qwen3report,llama3report,gemma2report,phi4report,phi4minireport,olmo2report,olmoereport,olmohybridreport};
Table~\ref{tab:models} (Appendix~\ref{app:tables}) lists them.
Six are post-trained (the four Qwen3 releases and both Phi-4 models); we report base-only
statistics over the other ten where post-training could matter. All run in bf16 on one 48\,GB
GPU. Prompts are completion-style, at 0 or 3 demonstrations. At zero shots no arm contains a
demonstration, which makes it the clean contrast; most claims are reported at both counts.

\paragraph{Prompt arms and positions.}
For every example we run three prompts on the same word and label:
the \textbf{question}, the task prompt (e.g.\ \texttt{The last letter of the word "cat" is});
a \textbf{neutral} sentence, \texttt{The word "cat" appears in this sentence.}; and a
\textbf{continuation} carrier, \texttt{Consider the word "cat". The answer is}, which matches the
question's ``awaiting an answer'' final position without naming the property. Term tasks carry
their query letter or position into both carriers. We read hidden states at two positions: the
\textbf{answer position} (the final prompt token) in every arm, and the \textbf{word position} in
the question and neutral arms, for the pure-word tasks, as either the word's last token or the mean
over all tokens overlapping the word.

\paragraph{Probe and estimator.}
A standardized multinomial logistic regression ($\ell_2 = 10^{-3}$) is fit at each of 12 evenly
spaced layers (including the embedding layer and the last layer), with three-fold cross-validation
grouped by query word, so no query word appears in both training and test labels. Folds are
cold-started. The reported accuracy uses \emph{held-out layer selection}: for each test fold, the
layer is chosen on the other folds. With this estimator, the peak-over-layers accuracy exceeds the
held-out-selected accuracy by 0.001--0.019 per task, with no positive dependence on model depth,
and all representation numbers lie within $\pm 0.021$ of an absolute-stride, warm-started
re-estimate (Appendix~\ref{app:estimator}).

\paragraph{Baselines.}
The \textbf{majority} baseline predicts the training fold's modal answer. The \textbf{nuisance
floor} is the best of several predictors that see only the word: its length, its position-free
letter set, the query letter where the task has one, and their combination, fit on the same folds.
Predictors that are sufficient for the label (length for \texttt{word\_length}, the letter set for
\texttt{count\_distinct}) are excluded. We report accuracy \emph{above the floor} unless stated
otherwise.

\paragraph{Readout.}
The model's own accuracy is the argmax of its next-token logits over the answer-token set (26
letters, yes/no, or ten digits). At zero shots, completion prompts are a degenerate format for
readout: larger models continue the sentence rather than answer, so readout claims use three shots.

\paragraph{Statistics.}
Scaling associations are Pearson correlations or OLS slopes against $\log_{10}$ parameters, with
two-sided permutation $p$-values from 100{,}000 permutations of model size. Because each model
contributes every arm, a question$\times$scale interaction equals the slope of the per-model
difference between arms. We also report a within-family (fixed-effect) slope, a 95\% interval
from resampling families, and base-only correlations. Where a claim spans the four
prompt-sensitive tasks we apply Holm correction \citep{holm1979}. The 16 checkpoints are related
and not independent draws; these are cross-model associations, not population scaling laws.

\section{A Prompt-Independent Word-Level Store}
\label{sec:store}

\begin{figure}[tb]
  \centering
  \includegraphics[width=\linewidth]{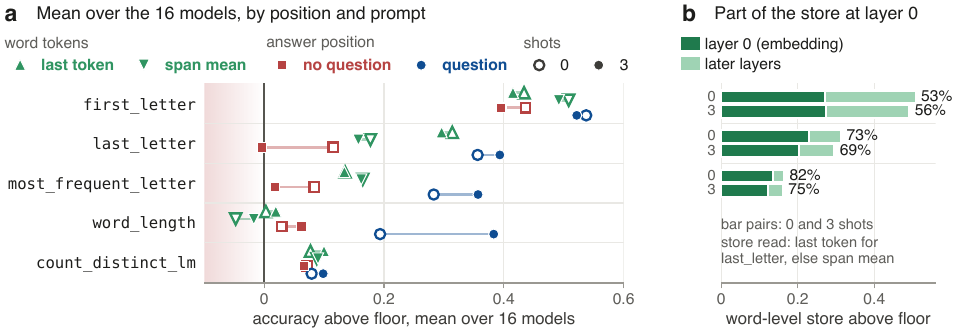}
  \caption{The word-level store. (a)~Accuracy above the word-only floor (held-out layer selection),
    mean over the 16 models: the query word's tokens in the neutral sentence (last token, or mean
    over the word's tokens) and the final prompt token without and with the question, at zero shots
    (hollow) and three shots (filled). For
    \texttt{count\_distinct\_lm} the floor equals the majority baseline. (b)~Part of the
    word-level store (better read per task) already present at layer 0, the token embedding, at zero
    and three shots; the percentage is that part's share of the store.}
  \label{fig:store}
\end{figure}

\paragraph{The store is present in every model.}
The word tokens encode \texttt{first\_letter}, \texttt{last\_letter} and
\texttt{most\_frequent\_letter} above the floor in all 16 models, at both shot counts and under
both reads (Figure~\ref{fig:store}a; exact values and model counts in Table~\ref{tab:store},
Appendix~\ref{app:tables}). Which read is stronger depends on where the property sits in the word: the last token for
\texttt{last\_letter} ($+0.314$ against $+0.177$ for the span mean at zero shots), the span mean
for \texttt{first\_letter} and \texttt{most\_frequent\_letter}.
We take the better read per task as the store in what follows; every conclusion below also holds
under the other read.

\paragraph{It is mostly in the embedding.}
At layer 0 (Figure~\ref{fig:store}b) the word tokens already carry 73\% of \texttt{last\_letter}'s above-floor store
(last-token read, zero shots; 69\% at three shots) and 82\% of \texttt{most\_frequent\_letter}'s
(span mean; 75\% at three shots). For \texttt{first\_letter} the share is 53--56\%, with the rest
added by intermediate layers. This agrees with direct evidence that token embeddings encode
spelling \citep{kaushal2022tokens,itzhak2022spelling}. Here it means that the prompt-independent
store is largely lexical, a property of the tokens that is read off with little contextual
processing.

\paragraph{It does not grow reliably with scale.}
Across the 16 models (Figures~\ref{fig:overview} and~\ref{fig:scaling}) the store's correlation with $\log_{10}$ parameters is $r = 0.47$
($p = 0.066$) for \texttt{last\_letter}, $r = -0.05$ for \texttt{most\_frequent\_letter} and
$r = -0.16$ for \texttt{first\_letter} (zero shots, last-token read; with the span mean, $r = 0.33$, $0.14$ and $-0.18$, none significant).
By contrast, the question arm's answer position correlates with scale at $r = 0.73$
($p = 0.002$) and $r = 0.72$ ($p = 0.0015$) for the first two tasks.

\paragraph{Where there is no store.}
\texttt{word\_length} is not linearly decodable at the word position above its floor, which here
is the letter-set predictor ($+0.003$ last token, $-0.048$ span mean at zero shots). Its answer
position under the question reaches $+0.383$ at three shots, so length, unlike the letter
properties, is assembled at the answer position rather than read off the word.
\texttt{count\_distinct\_lm} has a small store ($+0.08$ to $+0.10$ above majority in all 16 models)
that the question barely changes.

\paragraph{Sensitivity to the preceding context.}
The store is measured after the prefix \texttt{The word "}. In a pre-registered follow-up
(Appendix~\ref{app:prereg}) the prefix \texttt{In the word "} shifted the same word tokens' accuracy
by $+0.02$ to $+0.05$, with the word tokenized identically. Prefix sensitivity of this size is small
next to the store itself ($+0.13$ to $+0.51$ above the floor), but it sets the resolution at which
word-level differences between prompts can be interpreted.

\section{Asking Adds a Late Signal That Grows With Scale}
\label{sec:question}

\begin{figure}[tb]
  \centering
  \includegraphics[width=\linewidth]{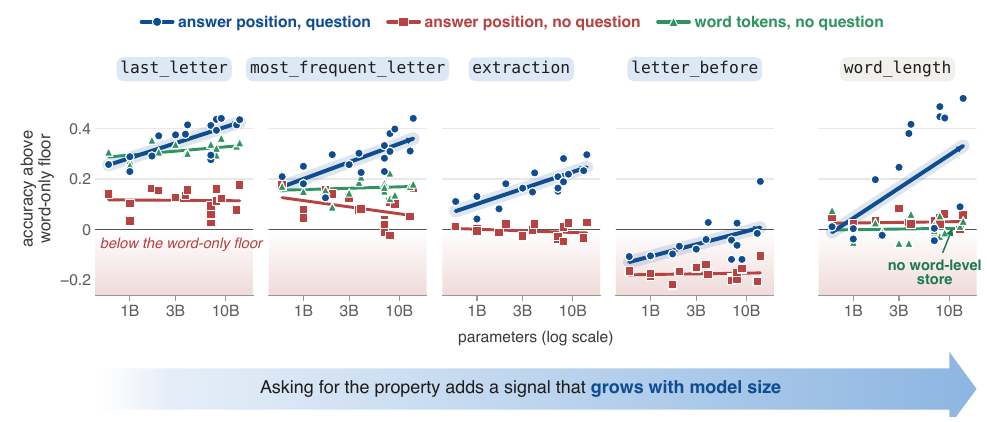}
  \caption{Accuracy above the word-only floor against model size, zero shots, one point per model
    with an OLS line per arm; the word tokens (triangles) are shown for the pure-word tasks
    (last-token read for \texttt{last\_letter} and \texttt{word\_length}, span mean for
    \texttt{most\_frequent\_letter}). Only the question arm rises with scale. The right panel
    contrasts a task with no word-level store: \texttt{word\_length} is assembled at the answer
    position (\S\ref{sec:store}). The red shading marks values below the floor. All values are in
    Table~\ref{tab:store} and the released analysis files.}
  \label{fig:scaling}
\end{figure}

\paragraph{The question effect grows with scale.}
Figure~\ref{fig:scaling} shows the four prompt-sensitive tasks, the tasks whose question-free
answer position retains little of the question arm's signal (below). Only the question arm rises
with scale; the neutral arm is flat or falling, and so is the word-level store. The interaction is
the slope of the per-model difference between the question and neutral arms
(Figure~\ref{fig:twophase}b; Table~\ref{tab:interaction} in Appendix~\ref{app:tables}). It is positive for all four tasks at both shot counts, survives Holm
correction across the four ($p_{\text{Holm}} \leq 0.018$), has a within-family slope of the same
size, and has a family-resampling interval that excludes zero.

\paragraph{The growth is late.}
At zero shots the question arm's layer profile has two parts (Figure~\ref{fig:twophase}a). In the
first quarter of the network it shows a bump at about 10\% depth that is similar in every model:
for \texttt{last\_letter} the early-band accuracy averages 0.75 and does not vary with scale
(slope $-0.002$ per decade, $p = 0.96$). One plausible source is the word's spelling being carried
to the adjacent answer token. Past half depth the profile rises again, and much further in larger
models.
Restricting both arms to the late band leaves the interaction intact for every task
($p_{\text{Holm}} \leq 0.017$ at zero shots, $\leq 0.016$ at three); restricting them to the early
band removes it (all $|$slope$| \leq 0.015$, $p \geq 0.42$; Figure~\ref{fig:twophase}b). The band
split also addresses a confound: the question prompt's answer token sits next to the word, while
the neutral sentence ends several tokens later. Distance could inflate the question arm early, but
it is the same in every model, so it cannot produce a scale trend. Nor does a trend appear in the
layers where distance would act.

\begin{figure}[tb]
  \centering
  \includegraphics[width=\linewidth]{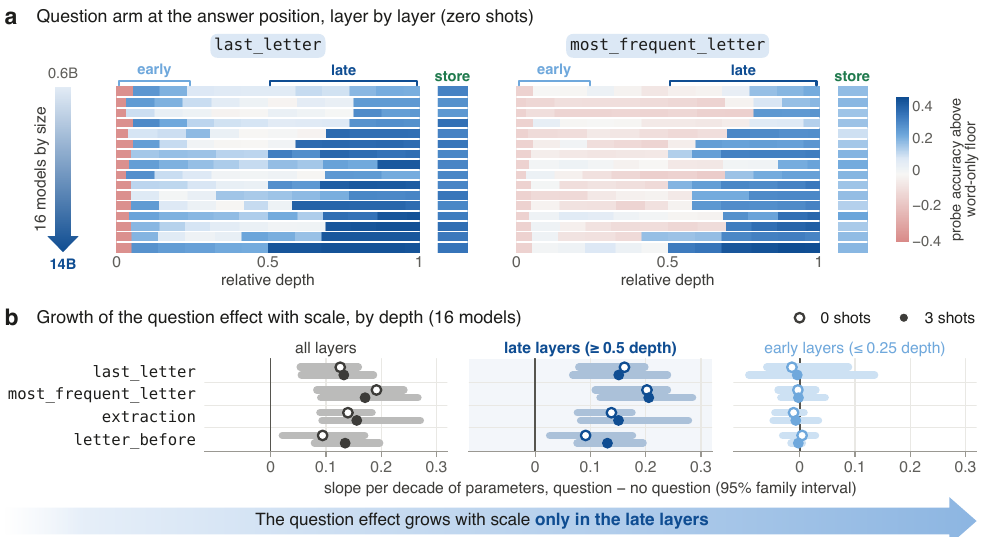}
  \caption{Depth and scale. (a)~The question arm at the answer position, zero shots: probe
    accuracy above the word-only floor at each probed layer, one row per model from smallest (top)
    to largest, each cell spanning that model's own relative depth. The separate column is the
    model's word-level store (word tokens in the neutral sentence; last token for
    \texttt{last\_letter}, span mean for \texttt{most\_frequent\_letter}). Brackets mark the early
    ($\leq 0.25$) and late ($\geq 0.5$) depth ranges. For \texttt{last\_letter} the early bump
    appears at every size, and the late band ends below the store in the smallest model and above
    it in the largest. (b)~Slope per decade of
    parameters of the question $-$ no-question difference at the answer position, zero shots
    (hollow) and three (filled), with 95\% family-resampling intervals, over all layers (held-out
    layer selection) or within one band. Slopes and Holm-corrected $p$-values are in
    Table~\ref{tab:interaction}.}
  \label{fig:twophase}
\end{figure}

\paragraph{Large models exceed their store; small models fall short of it.}
The answer position under the question does not simply relay the word-level store. For
\texttt{last\_letter} at zero shots, Qwen3-0.6B's answer position is below its own store
($+0.257$ against $+0.307$ above the floor), while phi-4's is above it ($+0.434$ against
$+0.344$). The difference (answer position under the question minus store) grows with scale for
both store tasks: $+0.091$ per decade ($p = 0.017$) at zero shots and $+0.098$ ($p = 0.0006$) at
three for \texttt{last\_letter}; $+0.129$ ($p = 0.011$) and $+0.113$ ($p = 0.0011$) for
\texttt{most\_frequent\_letter}. The same holds within the late band. In the larger models the
answer position therefore carries more linearly decodable information about the property than
the word's own tokens do. Whether that excess reflects computation, integration across the word's
tokens, or a change of format that a linear probe can exploit, our measurements do not say.

\paragraph{Where the late rise begins.}
As an exploratory description, the first depth past the early band at which the question arm
exceeds the store becomes shallower with scale for \texttt{most\_frequent\_letter} at three shots
($r = -0.76$, $p = 0.0009$; median relative depth 0.77 in the smaller eight models, 0.59 in the
larger eight). The other three task--shot cells lean the same way without reaching significance
($p = 0.07$--$0.31$), and 12 probed layers limit resolution to about 0.08 of depth, so we do not
treat this as a finding.

\begin{figure}[tb]
  \centering
  \includegraphics[width=\linewidth]{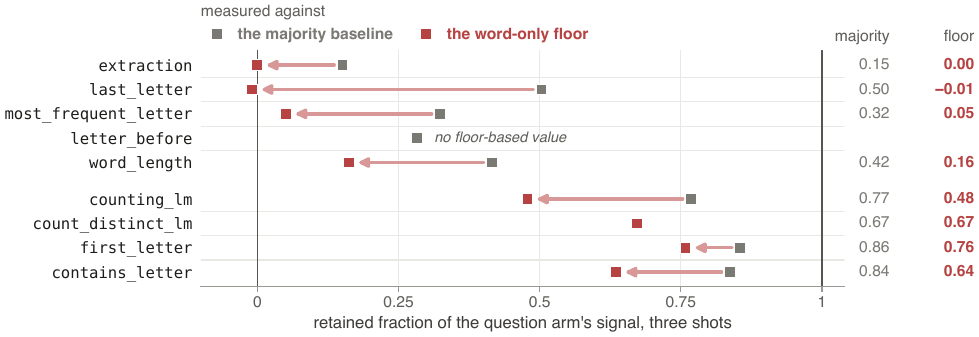}
  \caption{Retained fraction of the question arm's signal when the question is deleted, at three
    shots: mean(no-question $-$ base) / mean(question $-$ base) at the answer position of the
    neutral carrier, with the majority baseline (grey) or the word-only floor (red) as the base.
    Arrows run from the first to the second. \texttt{letter\_before} has only the majority-based
    value, because its question arm's mean is not above the floor. Zero-shot values and the
    continuation carrier are in Table~\ref{tab:retained}.}
  \label{fig:retained}
\end{figure}

\paragraph{What survives deleting the question.}
\emph{Retained} is the ratio of mean question-free accuracy to mean question accuracy, both measured
above a baseline (Figure~\ref{fig:retained}; Table~\ref{tab:retained}). Against the majority class,
\texttt{last\_letter} retains half its signal at three shots (0.503). Against the word-only floor it
retains none ($-0.009$, continuation carrier 0.034); \texttt{extraction} retains none and
\texttt{most\_frequent\_letter} 5\%. The majority-baseline figure was length and letter-set information that
any word-level predictor has. At zero shots more survives (\texttt{last\_letter} 0.32), consistent
with the early bump, which reaches the sentence-final token of the neutral carrier only when no
demonstrations intervene. \texttt{first\_letter} and \texttt{contains\_letter} retain 64--92\% and
are not question-driven; \texttt{word\_length} retains 15--16\% and joins the question-driven group.

\section{How Specific Is the Question Effect?}
\label{sec:crosstask}

To measure how much of the question effect of \S\ref{sec:question} depends on asking for
\emph{this} property, we probe label $A$ while the
prompt asks a different character question $B$ about the same word. The comparison is made at zero shots,
so the self and cross prompts differ only in the question sentence. Primary cells pair a letter-valued label
with a digit-valued question or the reverse, so $B$'s answer can never equal $A$'s label; cells
whose question arm is not at least 0.05 above the neutral arm are excluded. The two term-label tasks
(\texttt{extraction}, \texttt{letter\_before}) need their query term in every prompt, which we carry
as a bare note line (\texttt{Note the 3rd position.}); they form a secondary arm because the prompt
then differs in two elements.

\begin{figure}[tb]
  \centering
  \includegraphics[width=\linewidth]{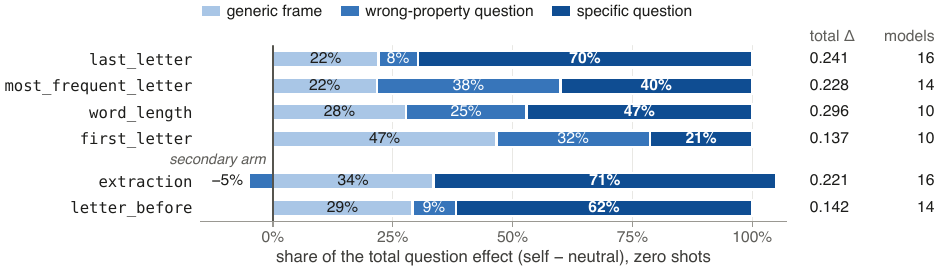}
  \caption{Decomposition of the question effect at zero shots, zero-leak cells, held-out layer
    selection: shares of the total $\Delta$ (self $-$ neutral, above-majority scale; right), with a
    negative share drawn left of zero. All 16 models were run; ``models'' counts those with at least one qualifying cell. Below the gap: the
    secondary term-label arm. The continuation floor is joined per (model, set, task) from the
    word-position runs, whose example draws differ from the cross-task runs'. Values and cell counts
    in Table~\ref{tab:crosstask}.}
  \label{fig:crosstask}
\end{figure}

We split the total question effect (self $-$ neutral) into three additive parts: the generic frame of
awaiting an answer (continuation $-$ neutral), a wrong-property character question
(cross $-$ continuation), and the specific question (self $-$ cross). Figure~\ref{fig:crosstask}
shows the shares.

For \texttt{last\_letter} and \texttt{extraction}, about 70\% of the effect requires asking for the
property itself; a wrong character question adds almost nothing ($8\%$ and $-5\%$).
\texttt{letter\_before} is similar (62\% specific, 9\% wrong-property). For
\texttt{most\_frequent\_letter}, a wrong character question already accounts for 38\% and the
specific question for 40\%, so a substantial part of its effect is shared by character questions in general.
\texttt{first\_letter}, whose answer position is near ceiling without any question
(\S\ref{sec:question}), is dominated by the generic frame. Preliminary runs with a peak-over-layers, warm-started estimator gave 70\%, 40\% and 72\% for
\texttt{last\_letter}, \texttt{most\_frequent\_letter} and \texttt{extraction}, within two percentage
points of the present values.

Every cross prompt is itself a character question about the same word, so this design measures how
specific the question effect is to the property asked. It does not separate a general character
routine from property-specific routing, and we draw no conclusion about mechanism from it. Each arm
is also read at its own held-out-selected layer, and the arms' layers differ in almost every cell,
so the shares split the best accuracy each prompt reaches, not accuracy at one fixed layer.

\section{Word-Only Floors and the Count Tasks}
\label{sec:nuisance}

\begin{figure}[tb]
  \centering
  \includegraphics[width=\linewidth]{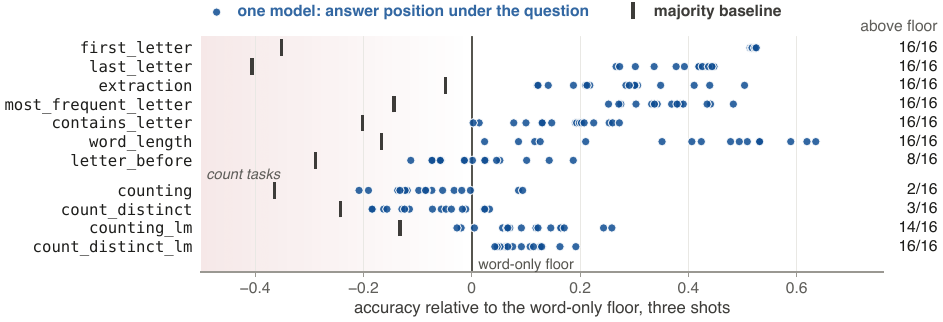}
  \caption{Word-only nuisance floors at three shots. Each dot is one model's question-arm probe
    (answer position, held-out layer selection), relative to the task's word-only floor (vertical
    line); the bar marks the majority baseline relative to the same floor (for
    \texttt{count\_distinct\_lm} the two coincide). Right: models above the floor. Below the gap:
    the original count tasks and their designed replacements. Floors, floor predictors and the best
    model per task are in Table~\ref{tab:nuisance}.}
  \label{fig:nuisance}
\end{figure}

A majority baseline controls for label skew, not for word properties that predict the label and
that the model encodes. Figure~\ref{fig:nuisance} shows, for every task at three shots, how each
model's question-arm probe compares with the word-only floor and the majority baseline
(Table~\ref{tab:nuisance} gives the values).

\paragraph{The original count tasks fail.}
\texttt{counting} clears its floor in 2 of 16 models and \texttt{count\_distinct} in 3. The cause is
our own sampling: stratifying \texttt{counting} by the true count made long words predict high
counts, because a letter occurring three times needs a long word. Length alone predicts the answer at
0.439 against a 0.311 majority. \texttt{count\_distinct} has the same problem, since the number of
distinct letters is bounded by word length (length alone: 0.532).

\paragraph{Length matching is not enough.}
We first rebuilt the count tasks with equal numbers of examples in every (word length, answer) cell,
which makes length statistically independent of the answer. For \texttt{counting} this left the floor
at 0.61, because two other shortcuts remained. The query letter alone predicted the count at 0.46
(about half of the three-occurrence items query \texttt{e}), and the number of repeated letters
(length minus distinct letters) predicted it at 0.59, since a word with no repeated letter can only
give a count of one.

\paragraph{Designed-out shortcuts make the task measurable.}
\texttt{counting\_lm} therefore also restricts words to exactly two repeated letters and draws the
query letter uniformly within each cell from letters that occur at every count at that length
(Appendix~\ref{app:lm}). Length, repetition and query-letter predictors then sit near the majority
baseline (0.25, 0.27 and 0.22 against 0.27), and the floor falls to 0.404, set by the letter set.
\texttt{counting\_lm} clears it in 14 of 16 models at three shots (16 of 16 at zero shots), by up to
$+0.259$. \texttt{count\_distinct\_lm} draws equal numbers per cell over lengths 8--9 and answers 5--8.
Its only informative word predictor, the letter set, determines the answer and is excluded, so its
floor equals the majority baseline and it tests above-majority decodability only.

\paragraph{The rebuilt count tasks are only partly question-driven.}
Without the question, \texttt{counting\_lm} keeps about half of its above-floor signal (0.48--0.52,
neutral carrier) and \texttt{count\_distinct\_lm} two thirds or more (0.67--0.91,
Table~\ref{tab:retained}). Neither shows a significant question$\times$scale interaction
(\texttt{counting\_lm}: $p = 0.24$ and $0.13$ at zero and three shots; \texttt{count\_distinct\_lm}:
$p = 0.97$ and $0.06$). Removing the shortcuts made the count tasks measurable, but it did not make
them further instances of the late, scale-dependent signal of \S\ref{sec:question}.

\paragraph{Other tasks.}
Six tasks clear their floors in every model by large margins. \texttt{letter\_before} clears its floor in 8 of
16 models at three shots and 3 of 16 at zero shots, and we treat it as marginal throughout.
\texttt{extraction}'s floor is low (0.171) because position-free letter information says little about
which position was asked.

\section{Readout: No Single Scaling Trend}
\label{sec:readout}

\begin{figure}[tb]
  \centering
  \includegraphics[width=\linewidth]{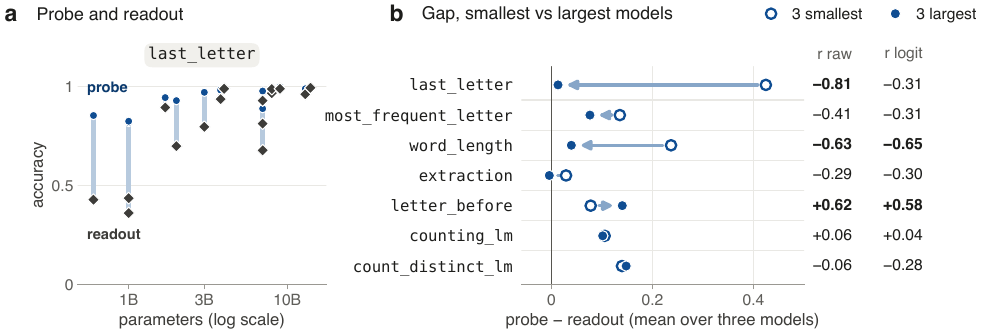}
  \caption{Readout at three shots. (a)~\texttt{last\_letter}: probe accuracy (held-out-selected) and
    the model's forced-choice readout accuracy, one pair per model, joined by the gap between them.
    (b)~The representation--readout gap (probe $-$ readout) per task, averaged over the three
    smallest and the three largest models, with its correlation with $\log_{10}$ parameters across
    all 16 models on the raw and the logit scale (bold: uncorrected permutation $p < 0.05$). After
    Holm correction across the $7 \times 3$ tests, only \texttt{last\_letter}'s raw-scale trend
    survives, and it is a ceiling effect that disappears on the logit scale
    (\S\ref{sec:readout}). Permutation $p$-values and base-only correlations are in
    Table~\ref{tab:gap}.}
  \label{fig:gap}
\end{figure}

The representation--readout gap is the held-out-selected probe accuracy minus the model's own
forced-choice accuracy, at three shots (Figure~\ref{fig:gap}; Table~\ref{tab:gap}).

\paragraph{The \texttt{last\_letter} narrowing is a ceiling effect.}
On the raw scale, \texttt{last\_letter}'s gap falls from $+0.424$ in the smallest models to $+0.013$ in
the largest ($r = -0.81$). In the largest models the probe is at 0.99--1.00 and readout catches up to
it. On the logit scale, which does not compress differences near 1.0, the association is
$r = -0.31$ ($p = 0.24$). Our pre-registered criterion required $|r| \geq 0.4$ \emph{and}
$p < 0.05$ on the logit scale to keep a narrowing claim, so we make none.

\paragraph{Other tasks disagree with one another.}
The gap narrows for \texttt{word\_length} on both scales ($r = -0.65$ on the logit scale,
$p = 0.005$), shows no trend for \texttt{extraction} (where it is near zero at every size), and grows
for \texttt{letter\_before} ($r = +0.62$, $p = 0.010$; base models only, $r = +0.80$). The two
rebuilt count tasks hold a gap of about $+0.10$ to $+0.15$ at every scale. After Holm correction
across the 21 tests of Table~\ref{tab:gap}, none of these trends survives (smallest adjusted
$p = 0.09$), and together they do not fit a single scaling trend. They also do not support the idea that
gaps close for retrieval-like tasks and persist for computational ones:
\texttt{extraction} and \texttt{letter\_before} are both retrieval-like and behave oppositely. The probe
also has hundreds of labeled examples while the model sees three demonstrations, so the gap is not
a calibrated measure of unused capacity. Readout accuracy also depends on task demands, which weigh
most on smaller models \citep{hu2024auxiliary}, so part of a small model's gap may reflect the
answer format rather than the representation.

\paragraph{The answer-token subspace.}
At the final layer the output head separates answers only along the span of the answer tokens'
unembedding rows. Projecting the final-layer state onto that span and re-probing beats a random
subspace of the same rank by $+0.12$ (160 of 176 cells, three shots) but beats the top principal
components of the same rank by only $+0.017$, and at zero shots not at all ($-0.002$). The span is
informative, but not much more so than a label-free comparator. It captures a median 7\% of the
between-class-mean energy. The loss from projection correlates with the readout gap ($r = 0.67$;
deficit 0.016, 0.051 and 0.098 across gap tertiles), but because the projected representation is
nested in the full one, this association does not show that the span causes the gap.

\paragraph{Post-trained models (exploratory).}
Six of the 16 models are post-trained. At three shots they lose less accuracy than the base models
when the final-layer state is projected onto the answer-token span (model-level means, difference
$-0.032$, permutation $p = 0.008$), and their readout is higher (0.62 against 0.52, $p = 0.049$).
Their span's margin over the PCA comparator is also larger, but not significantly ($+0.020$,
$p = 0.09$). Size does not explain these differences, since the post-trained models are slightly
smaller on average. At zero shots none of them is significant ($p \geq 0.09$). The pattern is
consistent with post-training placing more of the linearly decodable signal in the answer-token
span, but it is confounded with model family (the post-trained set is exactly the Qwen3 and Phi-4
models), and the test was not planned in advance. We report it as an observation, not a finding.

\section{Discussion}
\label{sec:discussion}

\paragraph{Two sources of character decodability.}
The data separate two sources of linearly decodable character information. One is stored with the
word: present at its tokens in a sentence that asks nothing, largely already in the token embedding,
and roughly constant across a more than 20-fold range of model size. The other appears at the answer position
only when the property is asked for, mostly in response to that specific property, and in the late
half of the network. Only the second grows with scale, so the common observation that larger models
``know'' more about spelling at the answer position is, in this battery, an observation about the
second source.

\paragraph{The late signal is not a relay of the store.}
If the question merely relayed the store (what a linear probe reads at the word's tokens), the answer
position should not systematically exceed it. It falls short of the store in small models and exceeds
it in large ones, and the difference grows with scale for both tasks where it can be measured. The excess may
come from combining information spread over the word's several tokens, from computation the
question triggers, or from re-encoding the property into a form a linear probe reads more easily.
One such form is the model's distribution over candidate answers, which probe directions at answer
positions have been reported to track \citep{yu2026answerbasin}; our answer-token-span test
(\S\ref{sec:readout}) bears on this reading only at the final layer. A
linear probe cannot tell the three apart, and we avoid the word ``routing'' because it implies the first
reading only.

\paragraph{The early phase.}
The early bump in the question arm is similar at every scale and is consistent with the word's
spelling being carried to the adjacent answer token by early attention. It explains why some
question-free signal survives at zero shots and why small models' held-out layer selection can land
in the first few layers. It is also a reason to report layer-restricted analyses: a peak over all
layers mixes an early signal that is similar at every scale with a scale-dependent late one.

\paragraph{What the controls changed.}
The controls are cheap to run, and each one changed a conclusion. Retained question-free signal
measured above the majority class was largely nuisance. Count tasks that appeared to show a
computation gap failed a word-only floor, and length matching alone did not fix them. A strong
raw-scale narrowing of the \texttt{last\_letter} readout gap disappeared on the logit scale. A
question placed before the word raised the word tokens' \texttt{last\_letter} decodability in all 16
models in our main runs, but the pre-registered follow-up shows that this is prefix sensitivity, not
anticipation: changing \texttt{The word "} to \texttt{In the word "} moves the same tokens by a
similar amount, and a question adds nothing measurable beyond that.

\section{Limitations}
\label{sec:limitations}

\paragraph{Probes are correlational.}
All claims concern linear decodability under a held-out-word probe. We do not show that the model
uses the store, the late signal or the answer-token span; interventions that pass modified states
through the model would be needed for that. A probe trained on hundreds of labels is not a calibrated
upper bound on what the model can do with three demonstrations, and we run no random-label
selectivity control.

\paragraph{Linear and positional scope.}
Absence of linear decodability at a position does not show absence of the information elsewhere or in
a nonlinear form. The word-position store is read from the last token or the mean of the word's
tokens; other reads could recover more. The store is measured after one prefix
(\texttt{The word "}), and word-token accuracy shifts by up to about 0.05 with the preceding context.
We do not separate single-token from multi-token query words, so we cannot say whether the answer
position's excess over the store in larger models requires combining information across a word's
tokens. Multi-token words are assembled into word-level representations at their last token in
early and middle layers \citep{kaplan2025tokens}, while information about their individual tokens
is erased there early \citep{feucht2024erasure}, so the last-token read of the store may behave
differently for single- and multi-token words. Folds hold out query words, not subword tokens: a
held-out word's last token can occur in training words, so part of the word-position store,
especially for \texttt{last\_letter} at layer 0, may reflect token identity rather than spelling
that generalizes across tokens. A split by token would separate the two. Nor do we fit nonlinear probes at the word position: a small MLP might read a larger store
than the linear probe does, in which case the late layers may partly make existing information
linearly readable rather than add to it.

\paragraph{Cross-model statistics.}
The 16 checkpoints come from seven related families with different data, base and post-trained
variants, and dense, hybrid and sparse architectures. They are not independent draws, and parameter
count is confounded with family and training choices. Within-family slopes and family resampling
reduce but do not remove this concern. Each cell uses one sampling seed; a three-seed check on five
models in preliminary runs found per-cell standard deviations of at most 0.042. Holm correction
covers the four-task claims; the descriptive readout correlations are reported uncorrected, with a
Holm check in \S\ref{sec:readout}. Held-out layer selection is not fully nested: the accuracies that
choose a layer for one test fold come from probes partly trained on that fold. We did not run a
nested selection; the largest peak-over-layers optimism, 0.019, indicates how far layer selection
can move these numbers.

\paragraph{Task and format scope.}
All tasks use English words of length 3--9 from one frequency-ranked vocabulary, single-token direct
answers and completion prompts at zero or three demonstrations. Chain-of-thought prompting
\citep{wei2022cot,kojima2022zeroshot}, character-separated
input, other scripts and other tokenizers may behave differently. \texttt{letter\_before} clears its
floor in only half the models, \texttt{counting\_lm} reuses words (405 unique words for 792 examples)
and \texttt{count\_distinct\_lm} tests only above-majority decodability.

\paragraph{Mechanism.}
We do not identify what produces the late signal or why it exceeds the store in large models.

\section{Conclusion}
\label{sec:conclusion}

Character properties of a word are linearly decodable from its own tokens in every model we tested,
largely from the token embedding and with little change across scale. Asking for a property adds a
second, late signal at the answer position, and only this second signal grows with model size; in
larger models it carries more linearly decodable information than the word's tokens themselves. Word-only
floors, designed-out shortcuts, held-out layer selection and pre-registered tests each removed a
claim that a simpler analysis would have made. Character-probing studies should report the
word-position baseline, word-only floors and layer-restricted analyses before interpreting a probe at
the answer position as knowledge the model has but does not express.

\bibliographystyle{plainnat}
\bibliography{references}

\appendix
\section{Methods}
\label{app:methods}

Experiments are run by \texttt{scripts/run\_v4\_suite.sh}, which calls \texttt{scripts/run\_grid.py}
(question arm, readout, accessibility), \texttt{scripts/run\_wordpos.py} (three prompt arms, word and
answer positions, and the stem-order follow-up) and \texttt{scripts/run\_crosstask.py}. Analyses are
\texttt{scripts/analyze\_v4.py} and \texttt{scripts/analyze\_v4\_followups.py}; every table and figure
is generated by \texttt{scripts/make\_paper\_v4.py}.

\subsection{Stimuli}
\label{app:data}
Words are lowercase, alphabetic, length 3--9, from a 6000-word frequency-ranked list built from the
Brown Corpus \citep{kucera1967brown}. Each task draws 800 examples with seed 42; a word unsuitable for
a task (a tie for \texttt{most\_frequent\_letter}, no unique preceding letter for
\texttt{letter\_before}) is redrawn. The sampler iterates letters in sorted order, so example sets do
not depend on the Python hash seed. Classes with fewer than five examples are dropped before probing.

\subsection{Prompts}
\label{app:prompts}
Question stems, followed by \texttt{k} demonstrations of the form \texttt{STEM ANSWER.} when shots $>0$:

\begingroup\sloppy
\begin{quote}\footnotesize
\texttt{The first letter of the word "cat" is}\\
\texttt{The last letter of the word "cat" is}\\
\texttt{The 2nd letter of the word "cat" is}\\
\texttt{Does the word "cat" contain the letter "z"? Answer yes or no:}\\
\texttt{In the word "cat", the letter that appears most often is}\\
\texttt{In the word "cat", the letter immediately before "t" is}\\
\texttt{The word "cat" has this many letters:}\\
\texttt{The word "cat" has this many different letters:}\\
\texttt{In the word "cat", the letter "a" appears this many times:}
\end{quote}
\endgroup

The length-matched tasks use the stems of their base tasks. For digit answers that tokenize as a space
followed by the digit, the stem ends with a space so that the digit is the next token. Carriers, with
the query term added for term tasks:

\begin{quote}\footnotesize
\emph{neutral}\quad\texttt{The word "cat" appears in this sentence.}\\
\phantom{\emph{neutral}}\quad\texttt{The word "cat" and the letter "a" appear in this sentence.}\\
\emph{continuation}\quad\texttt{Consider the word "cat". The answer is}
\end{quote}

At three shots each carrier is preceded by three neutral sentences about other words, which carry no
answers. The answer position is the last non-padding token, located correctly for either padding
side. The word position uses the tokenizer's character offsets for the last quoted occurrence of the
word: its last overlapping token, or the mean over all overlapping tokens.

\subsection{Length-matched count tasks}
\label{app:lm}
\textbf{\texttt{counting\_lm}}. Candidate items are (word, query letter) pairs whose word has exactly
two repeated letters (length minus number of distinct letters $= 2$) and whose count is 1, 2 or 3. At
each length, only query letters with at least two items at every count are kept. The design rectangle
is the widest contiguous range of lengths in which every (length, count) cell has at least 20 items:
lengths 6--9, counts 1--3. Each of the 12 cells receives $\lfloor 800/12 \rfloor = 66$ examples (792
in total); within a cell the query letter is drawn uniformly from the kept letters, then an item with
that letter. This yields 405 unique words.

\textbf{\texttt{count\_distinct\_lm}}. Words are binned by (length, number of distinct letters); the
rectangle with the most answer classes (then the most lengths) in which every cell has at least 20
words is lengths 8--9 and answers 5--8, with 100 examples per cell (674 unique words).

Word-only predictors at three shots for \texttt{counting\_lm}, against a 0.271 majority: length 0.251,
repetition (length minus distinct letters) 0.271, query letter 0.222, letter set 0.404. For the intermediate,
length-matched-only design they were 0.273, 0.594 and 0.46 for the first three, with a floor of 0.611.

\subsection{Probe and estimator}
\label{app:estimator}
The probe is multinomial logistic regression on standardized features, fit with L-BFGS
($\ell_2 = 10^{-3}$, at most 200 iterations) on the GPU. Folds are three groups of unique query words
assigned by a fixed permutation, and each fold's fit starts from zero (cold start). Twelve layers are
probed per model, evenly spaced from the embedding layer to the last layer. Held-out layer selection
chooses, for each test fold, the layer with the best mean accuracy on the other two folds, and
averages the resulting test accuracies.

\paragraph{Estimator check.}
Preliminary runs used the maximum over layers at an absolute stride of four (so deeper models had
more layers to choose from) and warm-started each fold from the previous fold's solution. Under the
present estimator, the per-task mean selection optimism (peak minus held-out-selected accuracy)
ranges from 0.001 to 0.019, and its correlation with model depth ranges from $-0.57$ to $+0.48$
across tasks, with most values negative. Representation accuracy at three shots differs from the
preliminary estimator's by $-0.021$ to $+0.004$ per task. Scaling results are therefore not an
artefact of layer selection or warm starts.

\subsection{Nuisance floors}
\label{app:nuis}
For each task and shot count we rebuild the exact example set and fit, on the same folds, predictors
that see only the word: length (per-length modal answer), the 26-dimensional binary letter set, the
letter set plus a one-hot query letter (term tasks), and letter set plus length (with the query letter
for term tasks), each a logistic regression on standardized features. The floor is the best
non-oracle predictor, or the majority baseline if none beats it. Oracles are length for
\texttt{word\_length} and the letter set (with or without length) for the two
\texttt{count\_distinct} tasks.

\subsection{Accessibility}
\label{app:access}
At the final layer, after the model's final normalization, we take the mean-centred unembedding rows
of the kept answer tokens, form an orthonormal basis $Q$ of their span (rank $k-1$ for $k$ classes),
and re-probe $\tilde h Q Q^\top$. Comparators are three random orthonormal subspaces of the same rank
and the top principal components of the same rank, computed without labels.

\subsection{Construction of Figure~\ref{fig:overview}}
\label{app:fig1}
Panel~(b) uses the \texttt{last\_letter} word-position runs at zero shots. Each mountain's crest is
that model's question-arm probe accuracy above the word-only floor at the answer position, at each
of the 12 probed layers, smoothed between them with a shape-preserving (PCHIP) interpolant that
does not overshoot; values below the floor are drawn at the floor. The width and shape of each
mountain across the model axis are decorative (a Gaussian profile), and neighbouring mountains are
combined by their maximum, so no point blends two models. The green rim and the tick at the end of
each mountain mark that model's store (last token, held-out layer selection); navy marks crest
above that store. The five models whose crest never rises above their store are Qwen3-0.6B,
Llama-3.2-1B, Qwen3-1.7B, OLMoE-1B-7B and OLMo-Hybrid-7B. Faint lines mark heights 0.1 and 0.2
above the floor, and because the projection is orthographic the scale bar applies to every
mountain. The bars in the right-hand column are each model's best layer past half depth,
averaged over the 16 models, with and without the question. The prompt cards in panel~(a) are
schematic. Figures~\ref{fig:scaling} and~\ref{fig:twophase} show the same values on conventional
axes.

\section{Pre-Registered Criteria and Outcomes}
\label{app:prereg}

Decision criteria were written into the project's analysis plan before the corresponding data
existed. All use held-out layer selection.

\subsection{Word-position experiment}
\label{app:prereg_e1}
\textbf{Criterion.} For \texttt{last\_letter}, \texttt{most\_frequent\_letter} and
\texttt{first\_letter} at zero shots: (i) the word-position store in the neutral sentence is above the
nuisance floor in at least 12 of 16 models; (ii) the slope of (answer position under the question
$-$ store) against $\log_{10}$ parameters is positive with permutation $p < 0.05$.
\textbf{Outcome.} (i) holds in 16 of 16 models for all three tasks under both word reads. (ii) holds
for \texttt{last\_letter} ($p = 0.009$ span mean, $0.017$ last token) and
\texttt{most\_frequent\_letter} ($p = 0.011$ and $0.010$). It fails for \texttt{first\_letter}
($p = 0.40$ and $0.49$), whose answer position is at ceiling in every model (above-floor accuracy
$+0.537$ against a maximum of $+0.543$).

\subsection{Secondary criteria}
\begin{itemize}
  \item \textbf{Readout gap.} Keep a narrowing claim for \texttt{last\_letter} only if the logit-scale
    correlation has $|r| \geq 0.4$ and $p < 0.05$. Outcome: $r = -0.31$, $p = 0.24$; claim dropped
    (\S\ref{sec:readout}).
  \item \textbf{Estimator.} Restate preliminary results if mean selection optimism exceeds 0.03 and
    correlates with depth at $r > 0.5$. Outcome: 0.001--0.019, correlations mostly negative; no
    restatement needed (Appendix~\ref{app:estimator}).
  \item \textbf{Count tasks.} Reinstate a count-task comparison if \texttt{counting\_lm} clears its
    floor in at least 8 of 16 models. Outcome: 14 of 16 at three shots, 16 of 16 at zero shots
    (\S\ref{sec:nuisance}).
  \item \textbf{Interaction.} Report the question$\times$scale slope for all four prompt-sensitive
    tasks whatever the result, and claim only those with $p < 0.05$. Outcome: all four, at both
    shot counts (Table~\ref{tab:interaction}).
\end{itemize}

\subsection{Stem-order follow-up (null result)}
\label{app:prereg_f3}
In the main runs, a question placed before the word (\texttt{The last letter of the word "cat" is})
raised the word tokens' \texttt{last\_letter} decodability relative to the neutral sentence in 16 of
16 models ($+0.062$). Prefixes differed between the two prompts, so we tested word order directly at
zero shots with three prompts per example:
\texttt{qfirst} (\texttt{The last letter of the word "cat" is}), \texttt{wfirst}
(\texttt{In the word "cat", the last letter is}) and the neutral sentence, with matching templates
for \texttt{first\_letter}, \texttt{most\_frequent\_letter} (``most frequent letter'') and
\texttt{word\_length} (``number of letters'').

\textbf{Criteria.} Primary: for \texttt{last\_letter} and \texttt{most\_frequent\_letter},
word@\texttt{qfirst} $-$ word@\texttt{wfirst} $> 0$ in at least 12 of 16 models with sign-test
$p < 0.05$ (last-token read). Prefix control: word@\texttt{wfirst} $-$ word@neutral has mean
$|\Delta| < 0.02$, since in \texttt{wfirst} the word precedes any question.

\textbf{Outcome: fail.} The primary prediction fails for \texttt{last\_letter} (11 of 16, $p = 0.21$)
and passes for \texttt{most\_frequent\_letter} (15 of 16, $p = 0.0005$), but the prefix control fails
for both ($+0.041$ and $+0.020$). The word is tokenized identically in all three prompts (checked with
the Qwen3 tokenizer), so under causal attention the word's tokens differ only in the preceding text:
\texttt{The last letter of the word "}, \texttt{In the word "}, or \texttt{The word "}. A change from
\texttt{The} to \texttt{In the} moves the word's encoding as much as the question does. No
\texttt{qfirst} advantage grows with scale.

\begin{table}[h]
  \centering\footnotesize
  \setlength{\tabcolsep}{5pt}
  \begin{tabular}{@{}llrrrrr@{}}
\toprule
task & read & \texttt{qfirst}$-$\texttt{wfirst} & positive & sign $p$ & \texttt{wfirst}$-$neutral & $r$ vs scale ($p$) \\
\midrule
\texttt{last\_letter} & last token & +0.021 & 11/16 & 0.210 & +0.041 & $-$0.12 (0.652) \\
\texttt{last\_letter} & span mean & +0.039 & 15/16 & 0.0005 & +0.051 & +0.07 (0.805) \\
\texttt{most\_frequent\_letter} & last token & +0.019 & 15/16 & 0.0005 & +0.020 & +0.20 (0.458) \\
\texttt{most\_frequent\_letter} & span mean & +0.013 & 11/16 & 0.118 & +0.015 & +0.07 (0.793) \\
\texttt{first\_letter} & last token & +0.007 & 11/16 & 0.057 & $-$0.014 & +0.32 (0.239) \\
\texttt{first\_letter} & span mean & +0.002 & 10/16 & 0.302 & +0.004 & +0.13 (0.620) \\
\texttt{word\_length} & last token & +0.089 & 16/16 & $<$0.0001 & +0.063 & +0.38 (0.147) \\
\texttt{word\_length} & span mean & +0.077 & 16/16 & $<$0.0001 & +0.078 & +0.40 (0.124) \\
\bottomrule
\end{tabular}
  \caption{Stem-order follow-up, zero shots, 16 models, held-out layer selection. Columns give
    the mean difference, models with a positive difference, a two-sided exact sign test, the prefix
    control (\texttt{wfirst} $-$ neutral), and the correlation of the difference with scale.}
  \label{tab:f3}
\end{table}

\section{Measurement Hygiene}
\label{app:hygiene}

Five measurement problems found during development shaped the present protocol.

\paragraph{Stimulus-pool size.}
On a pool of about 100 words, the same model and probe gave a memorization false positive under
random splits (\texttt{most\_frequent\_letter} at 1.000 for Qwen3-8B, peaking at layer 2) and an
underpowered false negative under held-out-word splits (0.358 against a 0.309 majority, with about 26
training words per fold). The 6000-word pool with held-out-word folds gives 0.845 against 0.256. The
regimes also differed in example count and layer stride, so this is a sensitivity demonstration
rather than a calibrated pool-size effect; it is why every result here uses the large pool.

\paragraph{Padding side.}
Answer positions were originally located as the attention-mask sum minus one, which is correct only
for right-padded batches. Gemma-2's tokenizer pads on the left, so its answer-position states were
read from the wrong token. An unbatched free-generation check exposed this: Gemma-2-9B scored 0.098
on \texttt{first\_letter} under batched scoring and 1.000 under free generation. The error grew
monotonically with shot count, as longer prompts mean more padding. The current code locates the last
real token for either padding side, and all Gemma-2 numbers here come from corrected runs.

\paragraph{Final normalization.}
In Hugging Face decoder models the last returned hidden state already includes the final
normalization. An earlier version of our readout code normalized it a second time before the output
head. Probes read hidden states directly and were unaffected, but the forced-choice answer changed in
a median 7\% of examples per model and task (27\% at the 90th percentile). All readout and
answer-token-subspace numbers here come from a corrected run that reads the model's own logits. On
letter tasks at three shots, the corrected readout tracks unbatched free generation slightly more
closely than the earlier one did ($r = 0.982$ against $0.978$ over 80 model--task cells). The subspace
analysis recomputes the logits from the final hidden state; this reproduces the model's own answer
in at least 99.8\% of examples for 12 models and in 90--99\% for Gemma-2, OLMo-2-13B and phi-4.
Dropping cells below 99\% agreement moves the three-shot subspace means by less than 0.003.

\paragraph{Free-generation scoring.}
The free-generation check counts a hit if the target starts or appears early in a short generation.
That is informative for 22--26-class letter answers but not for two- or three-class answers, where
almost any generation containing a digit scores. We therefore use free generation only as a
diagnostic for letter tasks and report forced-choice readout everywhere.

\paragraph{Sampler determinism.}
An initial version of the sampler iterated Python sets when choosing query letters, so four tasks depended on the
hash seed. The sampler now iterates sorted sets, and analysis code orders models by size and then
name, so all reported numbers are identical across hash seeds.

\section{Result Tables}
\label{app:tables}

Exact values behind the figures in the main text, generated from the analysis outputs by
\texttt{scripts/make\_paper\_v4.py}.

\begin{table}[!ht]
  \centering\footnotesize
  \setlength{\tabcolsep}{4pt}
  \begin{tabular}{@{}lll@{}}
    \toprule
    family & sizes & notes \\
    \midrule
    Qwen3 & 0.6B, 1.7B, 4B, 8B & post-trained releases \\
    Llama-3.x & 3.2-1B, 3.2-3B, 3.1-8B & base \\
    Gemma-2 & 2B, 9B & base; right padding enforced (Appendix~\ref{app:hygiene}) \\
    OLMo-2 & 1B, 7B, 13B & base \\
    Phi-4 & 3.8B-mini-instruct, 14B & post-trained \\
    OLMo-Hybrid & 7B & DeltaNet/attention hybrid \\
    OLMoE & 1B-7B & sparse MoE, plotted at 7B total parameters \\
    \bottomrule
  \end{tabular}
  \caption{The 16 models. Every experiment covers all 16.}
  \label{tab:models}
\end{table}

\begin{table}[!ht]
  \centering\footnotesize
  \setlength{\tabcolsep}{4pt}
  \resizebox{\textwidth}{!}{\begin{tabular}{@{}lrrrrrr@{}}
\toprule
task & shots & floor & word, last token & word, span mean & answer, no question & answer, question \\
\midrule
\texttt{first\_letter} & 0 & 0.457 & +0.434 (16) & +0.508 (16) & +0.436 (16) & +0.537 (16) \\
\texttt{first\_letter} & 3 & 0.475 & +0.415 (16) & +0.492 (16) & +0.396 (16) & +0.522 (16) \\
\addlinespace[2pt]
\texttt{last\_letter} & 0 & 0.547 & +0.314 (16) & +0.177 (16) & +0.115 (16) & +0.356 (16) \\
\texttt{last\_letter} & 3 & 0.552 & +0.296 (16) & +0.157 (16) & $-$0.004 (6) & +0.393 (16) \\
\addlinespace[2pt]
\texttt{most\_frequent\_letter} & 0 & 0.442 & +0.135 (16) & +0.165 (16) & +0.084 (14) & +0.283 (16) \\
\texttt{most\_frequent\_letter} & 3 & 0.400 & +0.134 (16) & +0.162 (16) & +0.018 (10) & +0.357 (16) \\
\addlinespace[2pt]
\texttt{word\_length} & 0 & 0.363 & +0.003 (7) & $-$0.048 (2) & +0.029 (14) & +0.193 (12) \\
\texttt{word\_length} & 3 & 0.351 & +0.019 (10) & $-$0.018 (4) & +0.062 (15) & +0.383 (16) \\
\addlinespace[2pt]
\texttt{count\_distinct\_lm} & 0 & 0.216 & +0.077 (16) & +0.088 (16) & +0.072 (16) & +0.079 (16) \\
\texttt{count\_distinct\_lm} & 3 & 0.222 & +0.100 (16) & +0.090 (16) & +0.066 (16) & +0.098 (16) \\
\bottomrule
\end{tabular}}
  \caption{Accuracy above the word-only nuisance floor (held-out layer selection), mean over the 16
    models, with the number of models above the floor in parentheses (Figure~\ref{fig:store}).
    ``Word'' columns probe the query word's tokens in the neutral sentence (last token, or mean over
    the word's tokens); ``answer'' columns probe the final prompt token. For
    \texttt{count\_distinct\_lm} the floor equals the majority baseline because the letter set,
    which determines the answer, is excluded as an oracle.}
  \label{tab:store}
\end{table}

\begin{table}[!ht]
  \centering\footnotesize
  \setlength{\tabcolsep}{3.2pt}
  \resizebox{\textwidth}{!}{\begin{tabular}{@{}lrrrrrrrrr@{}}
\toprule
 & & \multicolumn{4}{c}{all layers, held-out-selected} & \multicolumn{2}{c}{late ($\geq$0.5)} & \multicolumn{2}{c}{early ($\leq$0.25)} \\
\cmidrule(lr){3-6}\cmidrule(lr){7-8}\cmidrule(l){9-10}
task & shots & slope & $p$ (Holm) & within-fam. & family CI & slope & $p_{\text{Holm}}$ & slope & $p$ \\
\midrule
\texttt{last\_letter} & 0 & +0.126 & 0.0003 (0.0009) & +0.125 & [+0.053, +0.159] & +0.162 & 0.0003 & $-$0.014 & 0.690 \\
\texttt{most\_frequent\_letter} & 0 & +0.192 & 0.0004 (0.0009) & +0.173 & [+0.084, +0.241] & +0.202 & 0.0003 & $-$0.003 & 0.900 \\
\texttt{extraction} & 0 & +0.140 & $<$0.0001 ($<$0.0001) & +0.160 & [+0.089, +0.183] & +0.138 & $<$0.0001 & $-$0.011 & 0.426 \\
\texttt{letter\_before} & 0 & +0.094 & 0.017 (0.017) & +0.104 & [+0.021, +0.170] & +0.091 & 0.016 & +0.004 & 0.714 \\
\addlinespace[2pt]
\texttt{last\_letter} & 3 & +0.132 & 0.0005 (0.002) & +0.129 & [+0.056, +0.186] & +0.151 & 0.004 & $-$0.005 & 0.921 \\
\texttt{most\_frequent\_letter} & 3 & +0.171 & 0.001 (0.004) & +0.157 & [+0.091, +0.266] & +0.205 & 0.0008 & $-$0.002 & 0.948 \\
\texttt{extraction} & 3 & +0.156 & 0.011 (0.011) & +0.192 & [+0.093, +0.271] & +0.151 & 0.015 & $-$0.007 & 0.670 \\
\texttt{letter\_before} & 3 & +0.135 & 0.003 (0.006) & +0.145 & [+0.079, +0.197] & +0.131 & 0.003 & $-$0.003 & 0.841 \\
\bottomrule
\end{tabular}}
  \caption{Question$\times$scale interaction (Figure~\ref{fig:twophase}b): slope per decade of
    parameters of (question arm $-$ neutral arm) at the answer position. ``All layers'' uses
    held-out layer selection; the permutation $p$ is followed by its Holm-adjusted value across the
    four tasks. ``Late'' and ``early'' restrict each arm to layers at relative depth $\geq 0.5$ or
    $\leq 0.25$ (maximum over the restricted layers); the late $p$ is Holm-adjusted across the four
    tasks. Family CI: 95\% interval from resampling model families.}
  \label{tab:interaction}
\end{table}

\begin{table}[!ht]
  \centering\footnotesize
  \setlength{\tabcolsep}{5pt}
  \begin{tabular}{@{}lrrrrrr@{}}
\toprule
 & \multicolumn{3}{c}{shots $=0$} & \multicolumn{3}{c}{shots $=3$} \\
\cmidrule(lr){2-4}\cmidrule(l){5-7}
task & majority & neutral & continuation & majority & neutral & continuation \\
\midrule
\texttt{extraction} & 0.228 & $-$0.036 & 0.240 & 0.151 & $-$0.000 & 0.017 \\
\texttt{last\_letter} & 0.671 & 0.324 & 0.473 & 0.503 & $-$0.009 & 0.034 \\
\texttt{most\_frequent\_letter} & 0.542 & 0.295 & 0.467 & 0.324 & 0.051 & 0.104 \\
\texttt{letter\_before} & 0.485 & -- & -- & 0.283 & -- & -- \\
\texttt{word\_length} & 0.546 & 0.149 & 0.439 & 0.416 & 0.162 & 0.175 \\
\texttt{counting\_lm} & 0.698 & 0.520 & 0.859 & 0.768 & 0.479 & 0.783 \\
\texttt{count\_distinct\_lm} & 0.911 & 0.911 & 1.082 & 0.673 & 0.673 & 0.610 \\
\texttt{first\_letter} & 0.885 & 0.811 & 0.855 & 0.856 & 0.758 & 0.696 \\
\texttt{contains\_letter} & 0.966 & 0.919 & 1.233 & 0.837 & 0.635 & 0.866 \\
\bottomrule
\end{tabular}
  \caption{Retained fraction of the question arm's signal when the question is deleted
    (Figure~\ref{fig:retained}): mean(no-question $-$ base) / mean(question $-$ base).
    ``Majority'' uses the majority baseline and the neutral carrier; the other columns use the
    word-only floor. Missing values: the question arm's mean is not above the floor
    (\texttt{letter\_before}).}
  \label{tab:retained}
\end{table}

\begin{table}[!ht]
  \centering\footnotesize
  \setlength{\tabcolsep}{6pt}
  \begin{tabular}{@{}lrrrrrr@{}}
\toprule
probed label & total $\Delta$ & generic frame & wrong-property $Q$ & specific $Q$ & cells & models \\
\midrule
\texttt{last\_letter} & 0.241 & 22\% & 8\% & \textbf{70\%} & 32 & 16 \\
\texttt{most\_frequent\_letter} & 0.228 & 22\% & 38\% & \textbf{40\%} & 28 & 14 \\
\texttt{word\_length} & 0.296 & 28\% & 25\% & \textbf{47\%} & 30 & 10 \\
\texttt{first\_letter} & 0.137 & 47\% & 32\% & \textbf{21\%} & 20 & 10 \\
\addlinespace[2pt]
\texttt{extraction} & 0.221 & 34\% & $-$5\% & \textbf{71\%} & 32 & 16 \\
\texttt{letter\_before} & 0.142 & 29\% & 9\% & \textbf{62\%} & 28 & 14 \\
\bottomrule
\end{tabular}
  \caption{Decomposition of the question effect at zero shots (Figure~\ref{fig:crosstask}),
    zero-leak cells, all 16 models, held-out layer selection. Shares are of the total $\Delta$
    (self $-$ neutral, above-majority scale). Below the rule: the secondary term-label arm. The
    continuation floor is joined per (model, set, task) from the word-position runs, whose example
    draws differ from the cross-task runs'.}
  \label{tab:crosstask}
\end{table}

\begin{table}[!ht]
  \centering\footnotesize
  \setlength{\tabcolsep}{5pt}
  \begin{tabular}{@{}lrrlrr@{}}
\toprule
task & majority & floor & floor predictor & models above floor & best above floor \\
\midrule
\texttt{first\_letter} & 0.123 & 0.475 & length + letter set & 16/16 & +0.525 \\
\texttt{last\_letter} & 0.145 & 0.552 & length + letter set & 16/16 & +0.447 \\
\texttt{extraction} & 0.122 & 0.171 & length + letter set & 16/16 & +0.503 \\
\texttt{most\_frequent\_letter} & 0.256 & 0.400 & length + letter set & 16/16 & +0.483 \\
\texttt{contains\_letter} & 0.476 & 0.678 & length + letter set & 16/16 & +0.272 \\
\texttt{word\_length} & 0.185 & 0.351 & letter set & 16/16 & +0.635 \\
\texttt{letter\_before} & 0.128 & 0.416 & length + letter set & 8/16 & +0.187 \\
\addlinespace[2pt]
\texttt{counting} & 0.311 & 0.676 & length + letter set & 2/16 & +0.094 \\
\texttt{count\_distinct} & 0.289 & 0.532 & length & 3/16 & +0.033 \\
\texttt{counting\_lm} & 0.271 & 0.404 & letter set & 14/16 & +0.259 \\
\texttt{count\_distinct\_lm} & 0.222 & 0.222 & majority & 16/16 & +0.192 \\
\bottomrule
\end{tabular}
  \caption{Word-only nuisance floors at three shots and the number of models whose question-arm
    probe (answer position, held-out layer selection) exceeds the floor (Figure~\ref{fig:nuisance}).
    ``Best above floor'' is the maximum over the 16 models. Below the rule: the original count tasks
    and their designed replacements.}
  \label{tab:nuisance}
\end{table}

\begin{table}[!ht]
  \centering\footnotesize
  \setlength{\tabcolsep}{5pt}
  \begin{tabular}{@{}lrrrrr@{}}
\toprule
task & gap, 3 smallest & gap, 3 largest & $r$ (raw) & $r$ (logit) & $r$ (base only, raw) \\
\midrule
\texttt{last\_letter} & +0.424 & +0.013 & $-$0.81 (0.0003) & $-$0.31 (0.242) & $-$0.92 (0.0003) \\
\texttt{most\_frequent\_letter} & +0.135 & +0.076 & $-$0.41 (0.118) & $-$0.31 (0.249) & $-$0.07 (0.854) \\
\texttt{word\_length} & +0.237 & +0.039 & $-$0.63 (0.007) & $-$0.65 (0.005) & $-$0.47 (0.165) \\
\texttt{extraction} & +0.029 & $-$0.004 & $-$0.29 (0.277) & $-$0.30 (0.252) & $-$0.51 (0.142) \\
\texttt{letter\_before} & +0.078 & +0.140 & +0.62 (0.010) & +0.58 (0.019) & +0.80 (0.006) \\
\texttt{counting\_lm} & +0.105 & +0.101 & +0.06 (0.811) & +0.04 (0.869) & +0.20 (0.556) \\
\texttt{count\_distinct\_lm} & +0.139 & +0.148 & $-$0.06 (0.831) & $-$0.28 (0.308) & +0.34 (0.341) \\
\bottomrule
\end{tabular}
  \caption{Representation--readout gap at three shots (Figure~\ref{fig:gap}): mean over the three
    smallest (Qwen3-0.6B, Llama-3.2-1B, OLMo-2-1B) and the three largest (phi-4, OLMo-2-13B,
    Gemma-2-9B) models, and its correlation with $\log_{10}$ parameters (permutation $p$). The
    logit column uses $\mathrm{logit}(\text{probe}) - \mathrm{logit}(\text{readout})$, which removes
    the compression near 1.0. No correction for the $7 \times 3$ tests is applied.}
  \label{tab:gap}
\end{table}

\end{document}